\documentclass[runningheads]{llncs}

\usepackage{eccv}

\usepackage{eccvabbrv}

\usepackage{graphicx}
\usepackage{placeins}
\usepackage{booktabs}

\usepackage{multirow}

\usepackage{listings}
\usepackage{amsmath,amssymb}
\usepackage{hyperref}
\usepackage{url}
\hypersetup{
  hidelinks,
  pdftitle={Hyperspectral Trajectory Image for Multi-Month Trajectory Anomaly Detection},
  pdfauthor={Md Awsafur Rahman, Chandrakanth Gudavalli, Hardik Prajapati, B. S. Manjunath}
}

\lstdefinestyle{pseudopython}{
  language=Python,
  basicstyle=\small\ttfamily,
  keywordstyle=\bfseries\color{blue!70!black},
  commentstyle=\itshape\color{green!50!black},
  stringstyle=\color{red!60!black},
  numberstyle=\tiny\color{gray},
  numbers=left,
  numbersep=5pt,
  xleftmargin=12pt,
  frame=single,
  framerule=0.4pt,
  rulecolor=\color{gray!50},
  backgroundcolor=\color{gray!5},
  breaklines=true,
  showstringspaces=false,
  tabsize=2,
  morekeywords={def,for,in,if,else,return,None,True,False,and,or,not,class,self,assert},
  deletekeywords={input},
  literate={->}{$\rightarrow$}{2} {<-}{$\leftarrow$}{2},
  aboveskip=6pt,
  belowskip=6pt,
}

\usepackage{orcidlink}

\usepackage{xcolor}

\renewcommand{\arraystretch}{0.9}  %
\begin{document}

\title{Hyperspectral Trajectory Image for Multi-Month Trajectory Anomaly Detection} 

\titlerunning{Hyperspectral Trajectory Image}

\author{Md Awsafur Rahman \and
Chandrakanth Gudavalli \and \\
Hardik Prajapati\and
B.~S.~Manjunath}

\authorrunning{M.~A.~Rahman et al.}

\institute{UC Santa Barbara, USA\\
\email{\{awsaf, chandrakanth, hardik, manj\}@ucsb.edu}}

\maketitle

\begin{abstract}
    Trajectory anomaly detection underpins applications from fraud detection to urban mobility analysis. Dense GPS preserves fine-grained evidence such as abnormal speeds and short-duration events, but its quadratic cost makes multi-month analysis intractable; sparse stay-point methods scale by discarding that evidence and require a separate modeling regime. We argue that this bottleneck is unnecessary: dense and sparse trajectories share a natural two-dimensional cyclic structure along within-day and across-day axes. TITAnD (Trajectory Image Transformer for Anomaly Detection) is the first framework to cast both in a single representation, a Hyperspectral Trajectory Image (HTI), a day $\times$ time-of-day grid whose channels encode spatial, semantic, temporal, and kinematic information. Under this formulation, agent-level detection reduces to image classification and temporal localization to semantic segmentation. The Cyclic Factorized Transformer (CFT) models the two temporal axes directly, reducing attention cost by up to two orders of magnitude. With the HTI, multi-month dense anomaly detection becomes feasible for the first time, and CFT keeps it accurate and fast. Empirically, TITAnD matches or beats every evaluated baseline in AUC-PR on all four benchmarks. It surpasses vision models such as U-Net while using as few as 6.5M parameters, and it runs 11--75$\times$ faster than a capacity-matched flat Transformer on the same trajectory images.
    \end{abstract}

\section{Introduction}

\begin{figure}[t]
    \centering
    \makebox[\linewidth][c]{\includegraphics[width=1.20\linewidth]{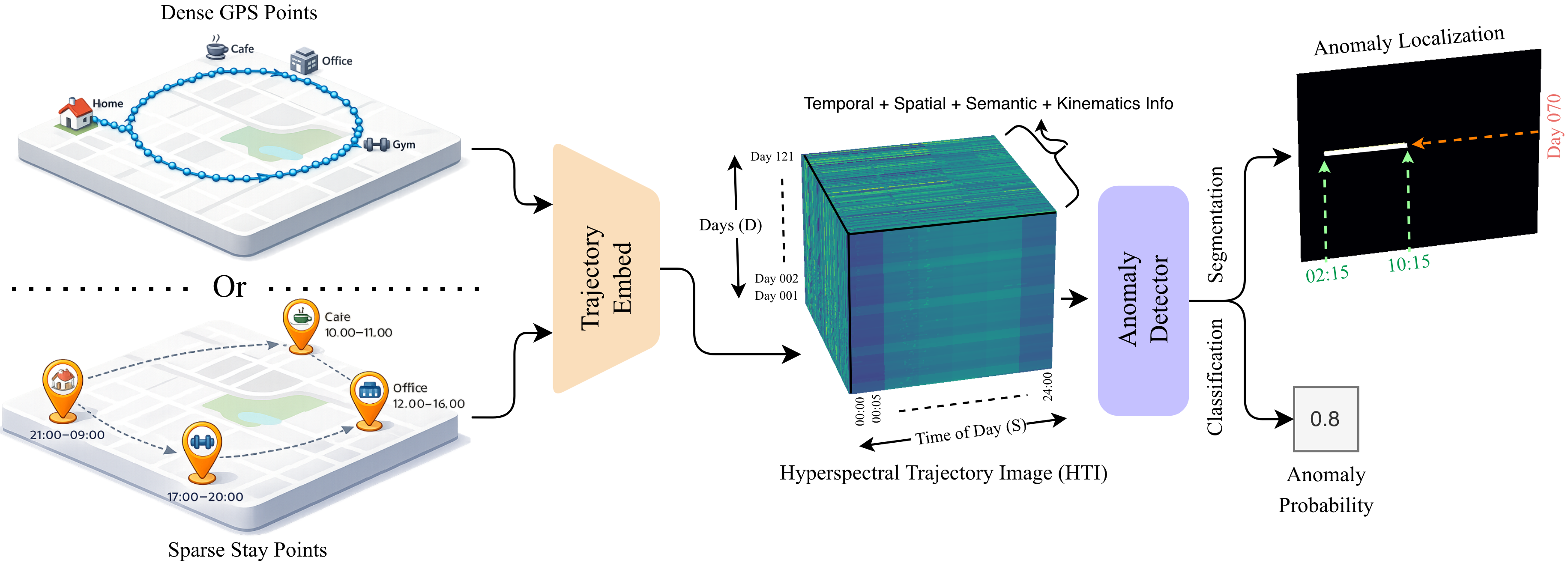}}
    \setlength{\abovecaptionskip}{4pt}
    \caption{Overview of our TITAnD framework. Dense GPS streams or sparse stay-points are encoded into a unified day $\times$ time-of-day hyperspectral trajectory image (HTI), where each pixel represents spatio-semantic, temporal, and kinematic information.}
    \label{fig:intro}
\end{figure}

Trajectory anomaly detection underpins applications from fraud detection and public safety to urban mobility analysis, where the goal is to identify rare or suspicious behaviors in an individual's movement history and localize when they occur~\cite{kennedy2024kinematic,azarijoo2025icad}.
Existing methods fall into two largely disconnected regimes. Sparse approaches compress dense GPS streams into event sequences such as stay-points or visits~\cite{zheng2009mining,wen2024ustad,azarijoo2025icad,hosseini2025semantic}, enabling long-horizon modeling but discarding short-duration activities and in-motion behaviors such as detours, abnormal speed profiles, and atypical routes~\cite{kennedy2024kinematic,fotraj}.
Dense approaches preserve this fine-grained evidence but face a long-context bottleneck: even after discretization, multi-month trajectories yield thousands of time steps, and Transformers scale quadratically with sequence length~\cite{vaswani2017attention,liang2022trajformer}.
To our knowledge, no existing method performs anomaly detection over multi-month dense GPS trajectories. Recent trajectory foundation models improve scalability through adaptive resampling or efficient encoders~\cite{zhu2024unitraj}, yet they still operate on flat one-dimensional sequences and do not target anomaly detection.
Likewise, prior work does not bridge dense and sparse regimes within one representation and architecture, and most existing approaches remain self-supervised, relying on reconstruction or next-event prediction as proxy objectives~\cite{wen2024ustad,azarijoo2025icad,rb_mbuya24}.
When labeled anomaly data is available, as in recent simulation benchmarks~\cite{shafqat2024numosim}, supervised detection remains largely unexplored.

Beyond the tension between observability and scalability, both paradigms share a deeper limitation: trajectories are modeled as flat one-dimensional sequences, ignoring the cyclic temporal structure governing human mobility~\cite{schneider2013unravelling}.
Within a day, movement unfolds as coherent episodes such as commutes, errands, or leisure, each characterized by consistent motion patterns and stop-and-go dynamics. 
Across days, behavior at the same time-of-day tends to follow stable routines, such as recurring morning commutes or evening gym visits~\cite{herder2012daily,schneider2013unravelling}. Anomalies, in turn, manifest as deviations along these two axes: an activity may be atypical for the time of day at which it occurs, or a behavior that recurs across days may be skipped or shifted.
Collapsing these temporal axes into a single dimension obscures this structure and forces models to rediscover periodicity from raw positional tokens.

This cyclic structure suggests a different formulation. If human mobility organizes along within-day and across-day axes, then trajectories over weeks or months naturally form an \emph{image}: a day $\times$ time-of-day grid where each pixel encodes behavior within a fixed time window (Fig.~\ref{fig:intro}). 
Under this formulation, agent-level anomaly detection reduces to image classification and temporal localization to semantic segmentation. 
Both dense GPS streams and sparse stay-point logs can be projected into the same image space through data-specific encoders that aggregate irregular observations into per-pixel embeddings, revealing that dense and sparse trajectory anomaly detection are instances of the same vision task. We instantiate this idea in \textbf{TITAnD} (\textbf{T}rajectory \textbf{I}mage \textbf{T}ransformer for \textbf{An}omaly \textbf{D}etection), an end-to-end supervised framework. Our contributions are as follows:

{\setlength{\leftmargini}{1.5em}
\begin{itemize}
  \item We present TITAnD, to our knowledge the first framework to perform anomaly detection over multi-month dense GPS trajectories. This is enabled by two complementary ideas: a Hyperspectral Trajectory Image (HTI) that compresses GPS observations into a structured 2D representation, and a Cyclic Factorized Transformer (CFT) that processes it efficiently.

  \item We unify dense and sparse trajectory data under a single architecture by reformulating anomaly detection as a vision problem with the HTI representation: a day $\times$ time-of-day grid whose channels encode spatial, semantic, temporal, and kinematic information aggregated from either dense GPS streams or sparse stay-points through modality-specific encoders.

  \item We introduce the Cyclic Factorized Transformer (CFT), which factorizes attention along the within-day and across-day axes of the HTI to exploit the cyclic pattern of human routines while reducing the cost significantly.

  \item Extensive experiments across dense and sparse benchmarks show that trajectory images alone do not guarantee improved performance: gains arise when the backbone explicitly exploits their cyclic structure.
\end{itemize}}

\section{Related Work}

\subsection{Trajectory Anomaly Detection}

Trajectory anomaly detection seeks to identify rare or suspicious behaviors in an individual's movement history. Existing approaches fall into two disconnected regimes by trajectory representation.

A common strategy is to compress dense GPS streams into visits or stay-points~\cite{zheng2009mining}. These representations scale well but discard short-duration and in-motion behavior. ICAD predicts visit sequences autoregressively~\cite{azarijoo2025icad}, USTAD reconstructs masked stay-point events~\cite{wen2024ustad}, and LM-TAD applies causal language modeling to user-conditioned trajectory tokens~\cite{rb_mbuya24}. Behavior-aware methods compare an individual's activity against their own normal profile~\cite{ai2024bestad,azarijoo2026ibad}, and CoBAD looks for anomalies in the joint behavior of related individuals~\cite{wen2025cobad}. These methods operate on sparse sequences and are typically self-supervised; we study supervised detection and evaluate all baselines on the same data, labels, and metrics.

When operating directly on dense GPS coordinates, existing work focuses on short horizons such as individual trips. FOTraj~\cite{fotraj} detects route anomalies in taxi trajectories using a spatio-temporal graph, while earlier methods such as iBAT~\cite{zhang2011ibat} identify deviations within individual rides. More broadly, recent trajectory architectures remain limited to short sequences: TrajFormer~\cite{liang2022trajformer} compresses attention for trip-level classification, TrajMamba~\cite{liu2025trajmamba} applies linear-time state-space models to vehicle trips, and UniTraj~\cite{zhu2024unitraj} pre-trains across large datasets but processes each trajectory as a flat one-dimensional sequence. Scaling these models to months-long histories would incur prohibitive quadratic cost or require local attention strategies that sacrifice cross-day context. Consequently, anomaly detection over dense, multi-month trajectories remains an open problem.

\subsection{From Cyclic Mobility Structure to Image Representations}

Human mobility exhibits strong regularity along two temporal dimensions. Within a day, movement follows circadian patterns with predictable episodes such as commutes, errands, and leisure~\cite{song2010limits,herder2012daily}. Across days, individuals repeatedly visit a small set of locations and follow stable routines over months~\cite{gonzalez2008understanding,schneider2013unravelling}. These regularities naturally define a two-dimensional temporal structure with days and time-of-day as separate axes, suggesting that long-horizon trajectories can be organized as a day $\times$ time-of-day grid rather than a flat sequence, as in eigenbehaviors~\cite{eagle2009eigenbehaviors}.

The idea of converting temporal signals into images has been explored in time-series analysis. Classical transforms such as Gramian Angular Fields, recurrence plots, and Markov transition fields encode temporal relationships as images for CNN-based classification~\cite{wang2015imaging,eckmann1987recurrence,hatami2018recurrence}. Other approaches render time series as visual representations processed by vision transformers~\cite{vitst}. These methods treat the image primarily as a visualization of a one-dimensional signal rather than a structured representation with semantically meaningful axes. Closer to our setting, TimesNet~\cite{wu2023timesnet} and the later TimeMixer++~\cite{timemixer++} reshape a time series along periods detected in the signal, using the two-dimensional view as a feature transformation within a general-purpose time-series backbone. In our work, the image is instead the formulation of the problem itself: it makes multi-month modeling efficient and organizes behavior along within-day and across-day axes so that our architecture can exploit the inherent two-dimensional structure of human mobility. Closest to us, Hosseini Sereshgi et al.~\cite{hosseini2025semantic} draw stay-points as a calendar image, color each visit by venue type, shade it by distance from the agent's usual places, and ask a vision-language model to spot anomalous agents without fine-tuning. Their image has no map location, nearby POIs, or motion, and it cannot hold dense GPS. In the HTI, every pixel carries learned spatial, semantic, temporal, and kinematic features for both dense GPS and stay-points, and CFT learns from it end to end.

\section{Method}
    \label{sec:method}

    \subsection{Overview and Task Definition}
    \label{sec:overview}
    
    TITAnD accepts two input modalities. A \emph{dense} trajectory is a time-ordered GPS stream
    $\mathcal{R}^{\text{dense}}=\{(\text{lat}_n,\text{lon}_n,t_n)\}_{n=1}^{N}$
    sampled at approximately 10-second resolution.
    A \emph{sparse} trajectory is a sequence of stay-point events
    $\mathcal{R}^{\text{sparse}}=\{(t^{\text{start}}_k,t^{\text{end}}_k,\text{lat}_k,\text{lon}_k)\}_{k=1}^{K},$
    each recording the arrival time, departure time, and location of a stationary episode.
    Both modalities cover multi-month observation windows with benchmark-specific past/future boundaries; PoL uses a 450-day history followed by a 14-day evaluation window.
    
    A data-specific encoder (Sec.~\ref{sec:tokenizer_method}) projects either input into a \emph{Hyperspectral Trajectory Image} (HTI), a $D\times S\times C$ tensor where $D$ is the number of days, $S$ is the number of time-of-day slots (288 five-minute slots for stay-point inputs and 96 fifteen-minute slots for dense GPS), and $C$ is the number of learned feature channels that encode temporal, spatial, semantic and kinematics information.

    Slot-level ground truth $y_{d,s}\in\{0,1\}$ is constructed from whichever label granularity is available.
    For dense trajectories with per-point annotations $\tilde{y}_n$, $y_{d,s}=\mathbb{I}\big[\max_{n\in\mathcal{R}^{\text{dense}}_{d,s}}\tilde{y}_n{=}1\big]$
    \label{eq:label_dense}, and for sparse trajectories with per-stop labels $\tilde{y}_k$, $y_{d,s}=\mathbb{I}\big[\exists\, k: \tilde{y}_k{=}1 \wedge [t^{\text{start}}_k,t^{\text{end}}_k)\cap\text{slot}(d,s)\neq\varnothing\big]$
    \label{eq:label_sparse}.
    Both constructions produce the same binary label grid $\mathbf{Y}\in\{0,1\}^{D\times S}$, enabling a unified loss regardless of input modality.

    \subsection{Trajectory Encoders}
    \label{sec:tokenizer_method}
    
    TITAnD provides two data-specific encoders---DenseTrajEmbed for dense GPS streams and SparseTrajEmbed for sparse stay-point logs---that each produce an HTI $\mathbf{X}\in\mathbb{R}^{D\times S\times C}$, where $C$ denotes the number of channels ($C{=}256$ in our experiments).
    Both encoders share three feature groups---spatio-semantic, temporal, and kinematic---but instantiate them differently for their respective input modality.
    The resulting HTI is consumed identically by CFT (Sec.~\ref{sec:cft_method}) regardless of data source.

    \begin{figure}[t]
        \centering
        \makebox[\linewidth][c]{\includegraphics[width=1.15\linewidth]{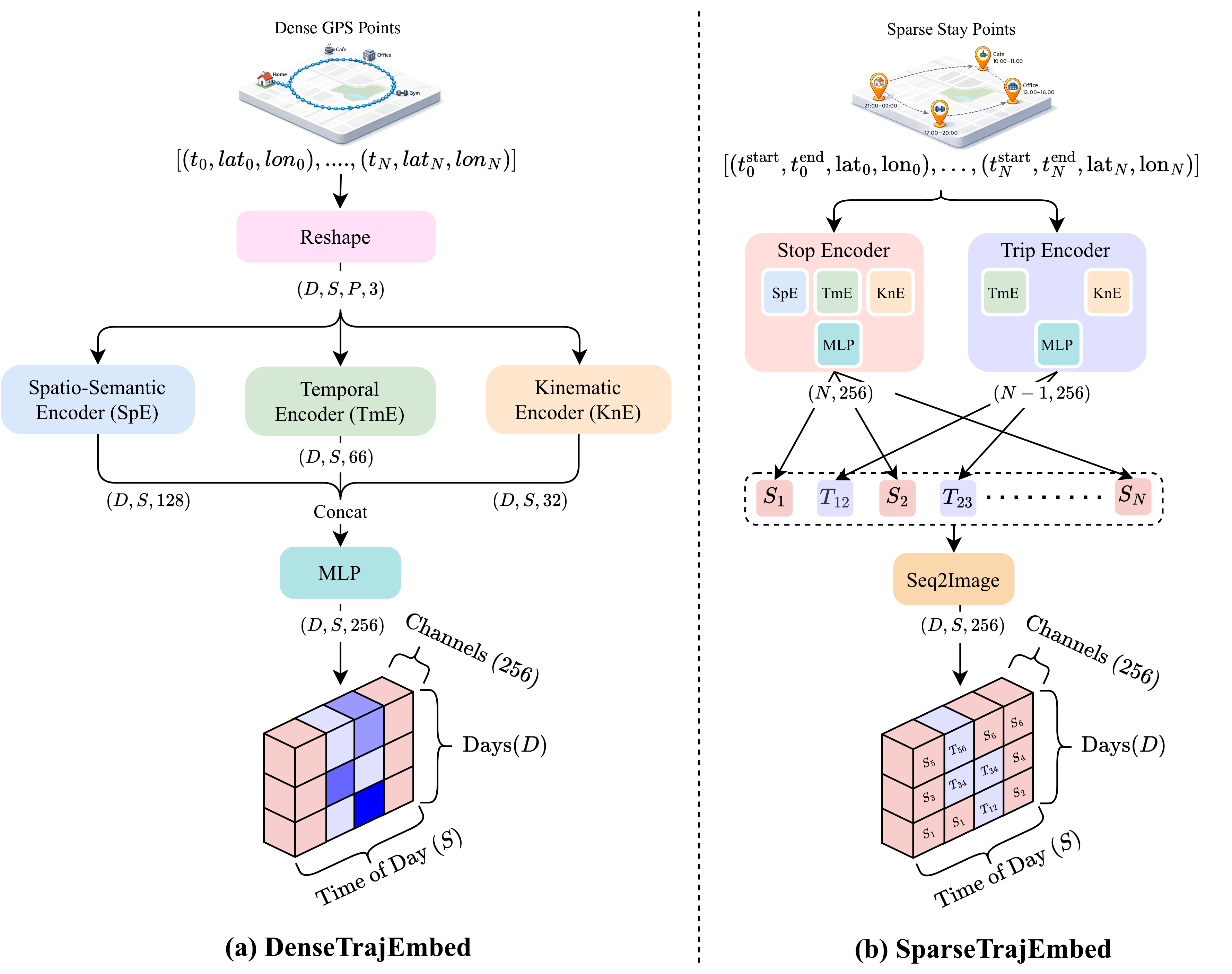}}
    \setlength{\abovecaptionskip}{4pt}
    \caption{TITAnD's two data-specific encoders. \textbf{(a) DenseTrajEmbed} fuses per-slot
    spatio-semantic (SpE), temporal (TmE), and kinematic (KnE) features from a GPS stream.
    \textbf{(b) SparseTrajEmbed} encodes stops with spatial, temporal, and kinematic features and trips with temporal and kinematic features, and
    Seq2Image maps each event onto its $(d,s)$ cells. Both produce the HTI consumed by CFT.}
        \label{fig:tokenizer}
    \end{figure}
    
    \subsubsection{Shared Components.}
    \label{sec:shared_components}
    
    \paragraph{Spatio-semantic cell embedding.}
    \label{sec:spatial_semantic}
    
    We first project GPS coordinates to a planar system (UTM) and assign each location to a grid cell. We use an adaptive QuadTree~\cite{samet1984quadtree} discretization that refines regions with high POI density, improving robustness to GPS jitter while keeping the vocabulary compact. Each location is mapped to a cell id $c$ and a discrete grid coordinate $(i,j)$.
    
    To inject lightweight semantics, each cell $c$ is associated with a set of nearby POIs $\mathcal{P}(c)$ (e.g. restaurant, office). It is capped to a fixed maximum per cell (125 for Tokyo and NumoSim-LA, 30 for PoL). For POI $q\in\mathcal{P}(c)$, we embed its category and relative position within the cell, $\mathbf{h}_{c,q}=f_{\text{poi}}\big(\text{cat}_{c,q}, \Delta x_{c,q}, \Delta y_{c,q}\big)\in\mathbb{R}^{64}$.
    We aggregate POIs into a compact cell summary via attention, $\mathbf{s}_c=\sum_{q\in\mathcal{P}(c)}\alpha_{c,q}\mathbf{h}_{c,q}$ with $\alpha_{c,q}=\mathrm{softmax}\big(\mathbf{w}^\top \mathbf{h}_{c,q}\big)$, so the model can emphasize functionally informative POIs when forming the cell semantics.
    We combine the cell semantics with a coarse absolute location code via sinusoidal positional encoding $\mathrm{PE}(i,j)\in\mathbb{R}^{128}$ to form the cell embedding $\mathbf{e}^{\text{cell}}_{c}\in\mathbb{R}^{128}$ (Eq.~\ref{eq:cell_emb}).
    This cell embedding module is shared by both dense and sparse encoders. Implementation details of QuadTree construction and POI buffering are deferred to supplementary material.
    
    \paragraph{Temporal encoder.}
    \label{sec:temporal_enc}
    Human mobility is highly periodic, so the same movement pattern can be typical or suspicious depending on when it occurs. We attach a single calendar embedding to each slot $(d,s)$ that captures daily, weekly, and longer-term context. Specifically, we encode a set of calendar attributes $\{\phi_k(d,s)\}_{k\in\mathcal{K}_{\text{time}}}$ and concatenate their encodings into $\mathbf{e}^{\text{tm}}_{d,s}\in\mathbb{R}^{66}$. Together with the cell embedding, this gives the two shared embeddings:
    \par\vspace{-0.5\baselineskip}\noindent
    \begin{minipage}{0.47\linewidth}
    \begin{equation}
    \mathbf{e}^{\text{cell}}_{c}=\mathrm{MLP}_{\text{sp}}\big([\mathbf{s}_{c};\mathrm{PE}(i_{c},j_{c})]\big),
    \label{eq:cell_emb}
    \end{equation}
    \end{minipage}\hfill
    \begin{minipage}{0.49\linewidth}
    \begin{equation}
    \mathbf{e}^{\text{tm}}_{d,s}=\mathrm{Concat}_{k\in\mathcal{K}_{\text{time}}}\big(\mathrm{Enc}_k(\phi_k(d,s))\big),
    \label{eq:time_emb_general}
    \end{equation}
    \end{minipage}
    \par\vspace{-0.3\baselineskip}\noindent where $\mathcal{K}_{\text{time}}$ includes hour-of-day, day-of-week, day-of-month, part-of-day (coarse bins), week-of-month, and month-of-year. This module is shared by both encoders.
    \vspace{-0.5em}

    \subsubsection{DenseTrajEmbed.}
    \label{sec:dense_traj_embed}
    
    DenseTrajEmbed (Fig.~\ref{fig:tokenizer}) converts a dense GPS stream into an HTI $\mathbf{X}\in\mathbb{R}^{D\times S\times C}$. The raw GPS stream is discretized into $D$ days and $S{=}96$ slots per day (15-minute slots; $D{=}66$ days in our experiments). Each slot holds $P{=}90$ points on a 10\,s grid, with gaps filled by same-day interpolation and zero padding applied at the end if necessary. Dense labels and encoder share the same 15-minute cells; 5-minute stay-point slots keep short stops and trips apart. This yields a regular tensor of shape $(D,S,P,3)$ together with a validity mask $m_{d,s,p}\in\{0,1\}$ for real values.
    
    \paragraph{Spatio-semantic pooling.}
    Each of the $P$ points in slot $(d,s)$ is mapped to a cell embedding $\mathbf{e}^{\text{cell}}_{c_{d,s,p}}$ via Eq.~\ref{eq:cell_emb}. We pool the $P$ point embeddings into a slot-level spatial token using learned attention pooling:
    \begin{align}
    \beta_{d,s,p} = \mathrm{softmax}\!\Big(\mathbf{v}_{\text{pool}}^\top \tanh(\mathbf{W}_{\text{pool}}\mathbf{e}^{\text{cell}}_{c_{d,s,p}})\Big),\quad
    \tilde{\mathbf{x}}^{\text{sp}}_{d,s} = \sum_{p=1}^{P}\beta_{d,s,p}\mathbf{e}^{\text{cell}}_{c_{d,s,p}}\in\mathbb{R}^{128}.
    \label{eq:pool_weight}
    \end{align}
    This produces one 128-d spatial token per slot, with the pooling weights allowing the model to emphasize informative points (e.g., transitions and turns) and downweight redundant samples.
    
    \paragraph{Kinematic encoder.}
    \label{sec:kinematic_enc}
    Many anomalies are expressed in motion rather than in endpoints, such as unusual speed profiles, abrupt accelerations, or atypical turning behavior. For each slot $(d,s)$, we summarize within-slot dynamics into a compact normalized descriptor vector $\mathbf{g}_{d,s}\in\mathbb{R}^{11}$ computed from consecutive GPS points inside the slot. It captures velocity and speed statistics, directionality (via $\sin/\cos$ of bearing), turning variability, acceleration, and extreme statistics (min/max speed and acceleration). We then map this descriptor to a slot-level embedding $\mathbf{e}^{\text{kn}}_{d,s}=\mathrm{Enc}_{\text{kn}}(\mathbf{g}_{d,s})\in\mathbb{R}^{32}$.
    
    \paragraph{Slot fusion.}
    \label{sec:slot_tokenization}
    The final slot token is obtained by fusing the pooled spatial token with slot-level temporal and kinematic context, $\mathbf{x}_{d,s}=\mathrm{MLP}_{\text{fuse}}\Big([\tilde{\mathbf{x}}^{\text{sp}}_{d,s}; \mathbf{e}^{\text{tm}}_{d,s}; \mathbf{e}^{\text{kn}}_{d,s}]\Big)\in\mathbb{R}^{C}$
.
    Stacking all slots yields the HTI $\mathbf{X}\in\mathbb{R}^{D\times S\times C}$.

    \subsubsection{SparseTrajEmbed.}
    \label{sec:sparse_traj_embed}
    
    SparseTrajEmbed converts sparse stay-point logs into the same $D\times S\times C$ HTI consumed by CFT. Instead of dense GPS points, the input consists of an interleaved sequence of stop and trip events: $S_1, T_{1 \rightarrow 2}, S_2, T_{2 \rightarrow 3}, \ldots$, where each stop $S_i$ records a location and dwell time, and each trip $T_{i \rightarrow i{+}1}$ records travel between consecutive stops.
    
    \paragraph{Event-to-grid mapping.}
    Each event spans a contiguous range of days and slots in the $D\times 288$ grid, determined by its start and end timestamps. Multi-slot events repeat their features across all occupied slots. When a stop and trip share the same 5-minute boundary slot, the slot keeps the stop's features, and a \emph{stop fraction} $f \in [0,1]$ records the stop's share of the slot minus the trip's share, clipped at zero.
    
    \paragraph{Three feature groups.}
    SparseTrajEmbed computes the same three feature groups as DenseTrajEmbed, instantiated differently for stops and trips.
    The spatio-semantic feature for stops is the cell embedding (Eq.~\ref{eq:cell_emb}) of the stop location; trip slots receive no spatio-semantic feature (zero vector).
    Attention along the slot axis lets a trip slot read its origin and destination from the stops around it, so trips need no location feature of their own.
    
The kinematic features differ by event type, $\mathbf{g}_\text{stop} = [\log_{10}(\text{dur}{+}1),\; 0]$ and $\mathbf{g}_\text{trip} = [\log_{10}(\text{dur}{+}1),\; \log_{10}(\text{dist}{+}1)]$,
where $\mathbf{g}_\text{stop}$ encodes only dwell duration (a zero distance signals a stationary event) and $\mathbf{g}_\text{trip}$ captures the travel time and straight-line distance between consecutive stops. Both are projected through a shared $\mathrm{Enc}'_{\text{kn}}$ to $\mathbb{R}^{32}$, analogous to $\mathrm{Enc}_{\text{kn}}$ in DenseTrajEmbed but with a 2-dimensional input suited to event-level kinematics rather than per-slot GPS statistics.
    
    \paragraph{Slot fusion and type embedding.}
    The three feature groups are concatenated and projected by a linear layer with layer normalization, $\mathbf{x}_{d,s} = \mathrm{Proj}\big([\mathbf{e}^{\text{sp}}_{d,s};\; \mathbf{e}^{\text{tm}}_{d,s};\; \mathbf{e}^{\text{kn}}_{d,s}]\big) \in\mathbb{R}^{C}$.
    A learned type embedding is then added to each occupied slot to distinguish stops from trips, $\mathbf{x}_{d,s} \leftarrow \mathbf{x}_{d,s} + f \cdot \mathbf{e}_\text{stop} + (1{-}f) \cdot \mathbf{e}_\text{trip}$,
    where $\mathbf{e}_\text{stop}, \mathbf{e}_\text{trip}\in\mathbb{R}^{C}$ are learned embeddings and $f$ is the stop fraction (0 for pure trip slots).
    The number of days $D$ is fixed at 66 for the dense encoder but varies across agents in the sparse setting; days added as padding never produce a prediction or enter the loss.

    \subsection{Cyclic Factorized Transformer (CFT)}
    \label{sec:cft_method}
    
    \begin{figure*}[t]
        \centering
        \makebox[\linewidth][c]{\includegraphics[width=1.30\linewidth]{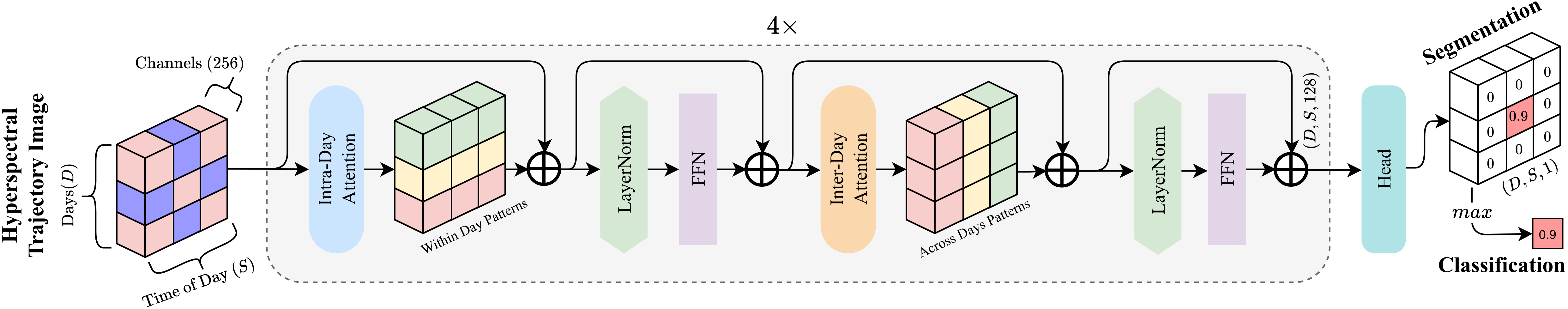}}
    \setlength{\abovecaptionskip}{4pt}
        \caption{Cyclic Factorized Transformer (CFT) factorizes attention by interleaving intra-day and inter-day attentions. Each intra-day attention captures within-day patterns, while each inter-day attention captures cross-day routine patterns.}
        \label{fig:cft}
    \end{figure*}

    \looseness=-1 Given the HTI $\mathbf{X}\in\mathbb{R}^{D\times S\times C}$ produced by either encoder, a naive approach is to flatten it into a length-$DS$ sequence and apply a standard Transformer. This is very expensive for multi-month horizons and, more importantly, it discards the temporal structure present in human mobility data. Specifically, human movement exhibits two complementary temporal structures: \emph{within-day episodes} (e.g., detours, stop-and-go transitions, prolonged dwells during a trip) and \emph{across-day routines} (e.g., recurring morning coffee, commute timing, regular gym visits). To preserve this cyclic day$\times$time-of-day structure while scaling to long horizons, we introduce the Cyclic Factorized Transformer (CFT). As shown in Fig.~\ref{fig:cft}, CFT factorizes attention along the two axes of the HTI $\mathbf{X}$, performing intra-day attention over the slot axis ($S$) to model daily episodes and inter-day attention over the day axis ($D$) to model routine consistency at the same time-of-day. This factorization reduces attention cost from $O(D^2S^2C)$ to $O((DS^2{+}SD^2)C)$, a $\frac{DS}{D{+}S}\approx 85\times$ reduction for $D{=}120$, $S{=}288$.

    CFT is related to axial attention~\cite{axial-attention,axial-deeplab} and to factorized video and multi-agent attention~\cite{bertasius2021timesformer,arnab2021vivit,ngiam2022scene}, but its axes are neither spatial nor agent dimensions: slots encode within-day episodes and days encode repeated routines. We therefore encode each token's slot and day with a rotary position encoding~\cite{su2024roformer}. As with the position embedding of ViT~\cite{dosovitskiy2020image}, it is applied once to the input tokens before the first CFT block, not to the queries and keys of every layer.

    \textbf{Intra-Day Attention.} Each intra-day layer applies multi-head self-attention over all $S$ slots within a single day, modeling short-horizon episode structure such as prolonged stays and unusual speeding. For a fixed day $d$, the row $\mathbf{X}_{d,:}\in\mathbb{R}^{S\times C}$ is treated as a length-$S$ sequence; collecting all days gives output in $\mathbb{R}^{D\times S\times C}$.

    \textbf{Inter-Day Attention.} Each inter-day layer mirrors the intra-day design but attends over the $D$ days at a fixed time-of-day slot $s$, capturing routine consistency and deviations across days such as changes in habitual commute or activity timing. The column $\mathbf{X}_{:,s}\in\mathbb{R}^{D\times C}$ is treated as a length-$D$ sequence; stacking across all slots yields output in $\mathbb{R}^{D\times S\times C}$.

    \textbf{Interleaved Blocks.}
    CFT stacks $L$ interleaved blocks (Fig.~\ref{fig:cft}; $L{=}4$ for stay-point inputs and $L{=}6$ for dense GPS), each applying IntraAttn $\to$ FFN $\to$ InterAttn $\to$ FFN with Pre-LN residual connections for stable optimization of the long temporal axes. Interleaving lets within-day evidence inform cross-day comparisons at every depth. The final representation $\mathbf{Z}=\mathbf{X}^{(L)}\in\mathbb{R}^{D\times S\times C}$ jointly encodes both scales.

    \subsection{Prediction Head}
    \label{sec:prediction}

    A two-layer MLP produces per-slot anomaly logits from the final representation: $\hat{y}_{d,s} = \sigma(\mathrm{MLP}_{\text{clf}}(\mathbf{Z}_{d,s}))$, yielding a segmentation map $\hat{\mathbf{Y}}\in[0,1]^{D\times S}$. Temporal localization reads directly from this map, while agent-level detection takes $\hat{y}_{\text{agent}}=\max_{d,s}\hat{y}_{d,s}$ over observed slots. We optimize $\mathcal{L}=\mathcal{L}_{\mathrm{BCE}}+\lambda\,\mathcal{L}_{\mathrm{Dice}}$ with a positive-class weight in the BCE term to address class imbalance; Appendix~\ref{sec:training_details} lists the per-benchmark values.

\section{Experiments}
\label{sec:exp}

\subsection{Experimental Setup}
\label{sec:exp_setup}

\textbf{Datasets.}
We evaluate TITAnD on four benchmarks (full statistics in Appendix~\ref{sec:datasets}). The public sparse benchmarks are NumoSim-LA~\cite{shafqat2024numosim} (about 80K agents, 55 days) and Urban Patterns of Life (PoL)~\cite{zufle2023urbanlife,amiri2023pol,rb_amiri24} (Atlanta, 3K agents, 464 days), which labels anomalous agents but not anomalous slots. No public dense trajectory anomaly benchmark exists, so Dense (Tokyo) uses private GPS from an agent-based simulator (about 18K agents, 66 days at 10\,s resolution), with anomalies injected by an independent red team whose rules were hidden from us. Dense2Sparse (Tokyo) uses a larger, distinct cohort from the same source converted to stay-points (about 80K agents, 120 days). All benchmarks use agent-disjoint splits.

\textbf{Evaluation and Metrics.}
We report area under the Precision--Recall curve (AUC-PR) and mean Intersection-over-Union (mIoU); mIoU is computed separately for each model and granularity at the F1-maximizing threshold on the evaluated predictions.
We evaluate at two granularities: \emph{temporal} (per-slot anomaly localization) and \emph{agent} (per-agent anomaly detection).
Sparse baselines (USTAD, ICAD) are originally self-supervised; for fair comparison we keep their published encoders and inputs but train them with the same supervision, using the same data, labels, and evaluation.

\textbf{Training Details.}
All benchmarks use the loss defined in Sec.~\ref{sec:prediction}. Sparse baselines replace their self-supervised heads with a classification head and are fine-tuned end-to-end without an auxiliary loss. Training uses mixed precision on NVIDIA H200 GPUs; Appendix~\ref{sec:training_details} gives the per-benchmark training settings.

\subsection{Comparison to Prior Methods}
\label{sec:exp_prior}

\textbf{Core comparison.}
Table~\ref{tab:main_results} reports AUC-PR and mIoU at temporal and agent granularity; PoL supplies agent-level metrics only. Sparse methods use stop or stay-point sequences, whereas the HTI backbones use the same trajectory images with a flat Transformer, U-Net, SegFormer, DINOv2+Adapter, or CFT.

Sparse methods such as USTAD~\cite{wen2024ustad} and ICAD~\cite{azarijoo2025icad} rely on stop-level representations. On NumoSim-LA, USTAD reaches 0.53 agent AUC-PR and ICAD reaches 0.30; at the temporal level they fall to 0.46 and 0.23, respectively. Both see the same stay-points as TITAnD, but as one long list of visits, where visits at the same time on other days can be far apart. The HTI puts them in one column, and inter-day attention compares them directly.

All HTI backbones use the same cell embedding, isolating the contribution of the backbone. The flat Transformer treats the day$\times$slot image as a one-dimensional sequence; U-Net~\cite{ronneberger2015u} uses an encoder--decoder CNN with skip connections; SegFormer~\cite{xie2021segformer} uses hierarchical attention with a lightweight decoder; and CFT factorizes attention along the day and slot axes. CFT is strongest or tied on every AUC-PR column: 0.59/0.65 temporal/agent AUC-PR on NumoSim-LA, 0.20/0.78 on Dense2Sparse Tokyo, and 0.84/1.00 on Dense Tokyo. The largest margin is temporal localization on dense GPS (0.84 versus 0.60 for the capacity-matched flat Transformer and 0.50 for U-Net), where global routine context and fine motion evidence both matter.

On the sparse benchmarks the flat Transformer is the weakest HTI backbone despite its more expensive full attention, suggesting that, even with calendar features, it must infer the cyclic day$\times$slot structure from positions in a long one-dimensional sequence. CNN backbones exploit a useful local inductive bias but have no explicit mechanism for comparing the same time of day across days. This distinction is clearest on Dense Tokyo: U-Net is close to CFT for coarse agent detection (0.95 versus 1.00 AUC-PR), which is near ceiling for most backbones, while CFT has its largest advantage on the harder temporal-localization task (0.84 versus 0.50).

\textbf{PoL, Chronos-2, and DINOv2.}
On PoL, TITAnD reaches 0.71 agent AUC-PR compared with 0.68 for SegFormer and 0.34 for LM-TAD~\cite{rb_mbuya24}, carrying the sparse comparison to a second public benchmark. On Dense Tokyo, TITAnD also exceeds both Chronos-2~\cite{rb_chronos2} and DINOv2+Adapter~\cite{oquab2024dinov2} in temporal AUC-PR.

\begin{figure*}[t]
    \setlength{\abovecaptionskip}{4pt}
    \centering
    \makebox[\linewidth][c]{\includegraphics[width=1.30\linewidth]{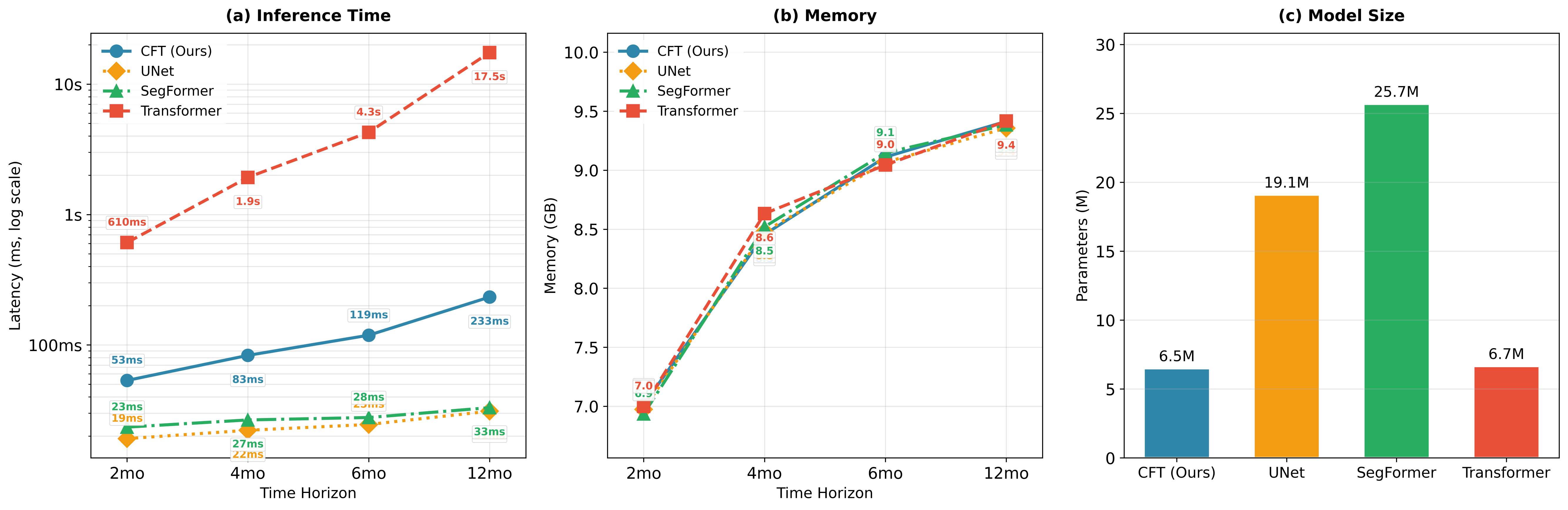}}
    \caption{Scaling over 2--12 months on one NVIDIA H200 (batch size 4) with sparse stay-point HTIs of shape $D\times288\times256$: (a) inference latency, (b) peak GPU memory, and (c) parameters.}
    \label{fig:scaling}
\end{figure*}

\textbf{Efficiency.}
CFT is the smallest evaluated HTI backbone, with 6.5M parameters on NumoSim-LA and PoL versus 19--26M for U-Net and SegFormer, and it scales favorably with the observation horizon (Fig.~\ref{fig:scaling}). At two months it is already 11$\times$ faster than the similarly sized flat Transformer (53\,ms versus 610\,ms); over 2--12 months, CFT's latency grows roughly linearly while the flat Transformer's grows quadratically, widening the gap to 75$\times$ at 12 months (233\,ms versus 17.5\,s). Peak memory remains comparable across backbones (about 7--10\,GB) with flash attention~\cite{dao2022flashattention}; U-Net and SegFormer are faster than CFT but less accurate (Table~\ref{tab:main_results}).

\begin{table*}[t]
    \centering
    \setlength{\abovecaptionskip}{4pt}
    \caption{Core comparison across dense and sparse benchmarks. Sparse methods use stop/stay-point sequences; HTI backbones use both input types. \texttt{---}: unsupported or unavailable.}
    \label{tab:main_results}
    \renewcommand{\arraystretch}{0.92}
    \setlength{\tabcolsep}{3pt}
    \resizebox{\textwidth}{!}{%
    \begin{tabular}{l l cc cc cc cc cc cc cc}
    \toprule
    & & \multicolumn{2}{c}{\textbf{PoL}} & \multicolumn{4}{c}{\textbf{Sparse (NumoSim-LA)}} & \multicolumn{4}{c}{\textbf{Dense2Sparse (Tokyo)}} & \multicolumn{4}{c}{\textbf{Dense (Tokyo)}} \\
    \cmidrule(lr){3-4} \cmidrule(lr){5-8} \cmidrule(lr){9-12} \cmidrule(lr){13-16}
    & & \multicolumn{2}{c}{Agent} & \multicolumn{2}{c}{Temporal} & \multicolumn{2}{c}{Agent} & \multicolumn{2}{c}{Temporal} & \multicolumn{2}{c}{Agent} & \multicolumn{2}{c}{Temporal} & \multicolumn{2}{c}{Agent} \\
    \cmidrule(lr){3-4} \cmidrule(lr){5-6} \cmidrule(lr){7-8} \cmidrule(lr){9-10} \cmidrule(lr){11-12} \cmidrule(lr){13-14} \cmidrule(lr){15-16}
    Type & Backbone & AUC-PR & mIoU & AUC-PR & mIoU & AUC-PR & mIoU & AUC-PR & mIoU & AUC-PR & mIoU & AUC-PR & mIoU & AUC-PR & mIoU \\
    \midrule
    \multirow{3}{*}{Sparse}
    & USTAD & 0.06 & 0.09 & 0.46 & 0.67 & 0.53 & 0.70 & 0.03 & 0.52 & 0.41 & 0.66 & --- & --- & --- & --- \\
    & ICAD & 0.06 & 0.36 & 0.23 & 0.59 & 0.30 & 0.62 & 0.01 & 0.51 & 0.36 & 0.65 & --- & --- & --- & --- \\
    & LM-TAD & 0.34 & 0.60 & 0.03 & 0.50 & 0.06 & 0.49 & --- & --- & --- & --- & --- & --- & --- & --- \\
    \midrule
    \multirow{5}{*}{HTI}
    & Transformer & 0.10 & 0.44 & 0.06 & 0.56 & 0.16 & 0.57 & 0.01 & 0.52 & 0.24 & 0.60 & 0.60 & 0.70 & \textbf{1.00} & \textbf{1.00} \\
    & U-Net & 0.44 & 0.69 & 0.29 & 0.63 & 0.57 & 0.75 & 0.03 & 0.52 & 0.60 & 0.72 & 0.50 & 0.67 & 0.95 & 0.77 \\
    & SegFormer & 0.68 & 0.76 & 0.33 & 0.66 & 0.62 & \textbf{0.76} & 0.03 & 0.53 & 0.52 & 0.70 & 0.21 & 0.59 & 0.95 & 0.91 \\
    & DINOv2+Adp. & 0.25 & 0.56 & 0.12 & 0.59 & 0.46 & 0.57 & 0.07 & 0.54 & 0.67 & 0.73 & 0.19 & 0.61 & \textbf{1.00} & 0.98 \\
    & \textbf{CFT (ours)} & \textbf{0.71} & \textbf{0.78} & \textbf{0.59} & \textbf{0.74} & \textbf{0.65} & \textbf{0.76} & \textbf{0.20} & \textbf{0.58} & \textbf{0.78} & \textbf{0.80} & \textbf{0.84} & \textbf{0.84} & \textbf{1.00} & \textbf{1.00} \\
    \midrule
    Time series & Chronos-2 & --- & --- & --- & --- & --- & --- & --- & --- & --- & --- & 0.69 & 0.74 & 0.82 & 0.61 \\
    \bottomrule
    \end{tabular}
    }%
\end{table*}

\subsection{Ablations}
\label{sec:exp_ablation}

\begin{table}[t]
    \setlength{\abovecaptionskip}{4pt}
    \caption{Ablation of encoder features and attention axes with temporal AUC-PR.}
    \label{tab:ablation}
    \centering
    \setlength{\tabcolsep}{6pt}
    \scriptsize
    \vspace{2pt}
    {%
    \begin{tabular}{l | c|c | c|c | c}
    \toprule
     & \multicolumn{2}{c|}{Encoder features} & \multicolumn{2}{c|}{Attention axes} & \\
    \cmidrule(lr){2-3} \cmidrule(lr){4-5}
     & Temporal & Spatial + Temporal  & Intra-day & Inter-day & \textbf{Full (CFT)} \\
    \midrule
    Dense (Tokyo) & 0.03 & 0.62 & 0.66 & 0.33 & \textbf{0.84} \\
    
     Dense2Sparse (Tokyo) & 0.18 & 0.19 & 0.12 & 0.07 & \textbf{0.20} \\
    
    \bottomrule
    \end{tabular}}
    \end{table}

We ablate both encoder features and attention axes while keeping the other component fixed (Table~\ref{tab:ablation}); a single-axis model doubles its layers to keep the parameter count.

\textbf{Encoder features.} On Dense, temporal features alone are insufficient (0.03), and adding spatial features yields the largest jump (0.03$\to$0.62), confirming that location context is the dominant signal in dense GPS. On Dense2Sparse, each feature group adds to the previous one (0.18$\to$0.19$\to$0.20). Kinematic features give a further gain on Dense (0.62$\to$0.84), reflecting that dense GPS retains rich motion dynamics.

\textbf{Attention axes.} On Dense (Tokyo), intra-day attention alone achieves 79\% of full AUC-PR, indicating that most anomaly signal in the dataset arises from within-day deviations. Inter-day attention alone captures 39\%, reflecting that cross-day comparison provides weaker signal when anomalies are locally coherent within individual days. On Dense2Sparse, intra-day and inter-day attention alone reach 0.12 and 0.07, below their factorized combination (0.20). Full interleaved attention is best among the attention variants, and the Full~(CFT) entries in both parts of Table~\ref{tab:ablation} correspond to the same model architecture.

\begin{figure}[t]
    \centering
    \makebox[\linewidth][c]{\includegraphics[width=1.20\linewidth]{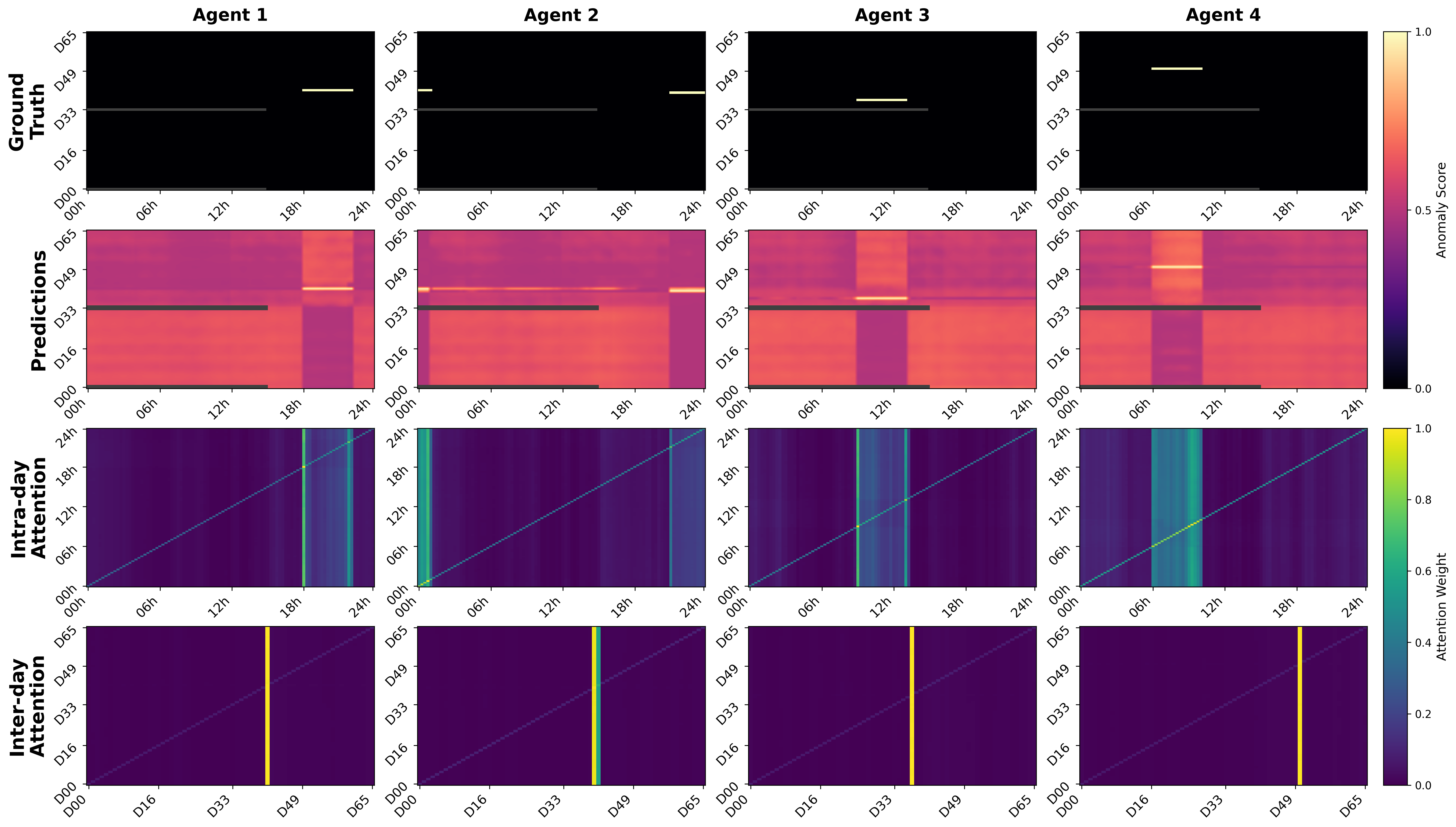}}
    \vspace{-4pt}
    \caption{Qualitative analysis. Rows show, from top to bottom, ground truth (gray denotes missing data), CFT predictions, intra-day (slot$\times$slot) attention, and inter-day (day$\times$day) attention.}
    \label{fig:qualitative}
\end{figure}

\paragraph{Qualitative analysis.}
Attention-rollout visualizations in Figure~\ref{fig:qualitative} show complementary localization along the two CFT axes: inter-day attention concentrates on anomalous days, while intra-day attention concentrates on the corresponding within-day slots. The same behavior appears for anomalies driven by wide-area movement, atypical dwelling, and prolonged inactivity.

\FloatBarrier
\section{Limitations}
\label{sec:limitations}
Because verified real-world mobility anomaly labels are scarce, our evaluation uses synthetically injected anomalies. Constructing verified labels requires access to sensitive mobility traces and ground-truth incident verification, which makes large public benchmarks difficult to obtain. Synthetic injection therefore provides a controlled way to compare detection performance across diverse anomaly types. For dense trajectories, no public dataset currently provides multi-month GPS traces with anomaly annotations, so our Dense (Tokyo) experiments use private data from an agent-based simulator; an independent red team injected anomalies using rules withheld from the modeling team. Consequently, two of our four benchmarks share the same private Tokyo source, which limits the geographic and demographic diversity of the evaluation. Results on additional independent real-world datasets would strengthen the generalization claims. Finally, USTAD and ICAD were originally designed for self-supervised learning. We adapt their encoders to our supervised protocol to compare representations under the same labels and evaluation setting, while recognizing that this differs from their original training objectives.

\section{Conclusion}
We showed that sparse and dense trajectory anomaly detection are instances of the same vision problem. Through hyperspectral trajectory images and structure-aware attention along the cyclic axes of human mobility, TITAnD makes multi-month dense anomaly detection feasible for the first time and unifies dense and sparse trajectories in a single representation. TITAnD matches or beats every evaluated baseline in AUC-PR on all four benchmarks, at as few as 6.5M parameters and up to 75$\times$ lower latency than a flat Transformer at 12 months. Our ablations show that the two attention axes complement each other: neither alone matches their factorized combination, on dense and sparse inputs alike.

\subsection*{AI use statement}
Generative AI tools were used in a limited assistive role for proofreading and code assistance.

\subsection*{Ethics statement}
Human mobility traces are sensitive and can reveal habits, associations, and frequently visited locations. Trajectory anomaly systems could cause privacy harm or be misused for surveillance if deployed without appropriate safeguards. We therefore advocate data minimization, access controls, retention limits, human oversight, and explicit documentation of dataset provenance, labeling assumptions, intended uses, and known limitations. Our experiments study research benchmarks and do not claim that model predictions alone are suitable for consequential decisions about individuals.

\subsection*{Reproducibility statement}
The appendix reports dataset statistics, preprocessing and label aggregation, pseudocode for HTI construction and CFT, encoder dimensions, training hyperparameters, and robustness analyses. The public NumoSim and PoL benchmarks support independent evaluation of the sparse setting. The private dense benchmark cannot be redistributed; we plan to release the implementation, preprocessing pipeline, and trained model artifacts so that the method can be evaluated on comparable authorized datasets.

\subsubsection*{Acknowledgements}
This work is supported by the Intelligence Advanced Research Projects Activity (IARPA) via Department of Interior/Interior Business Center (DOI/IBC) contract number 140D0423C0057. The U.S. Government is authorized to reproduce and distribute reprints for Governmental purposes notwithstanding any copyright annotation thereon. Disclaimer: The views and conclusions contained herein are those of the authors and should not be interpreted as necessarily representing the official policies or endorsements, either expressed or implied, of IARPA, DOI/IBC, or the U.S. Government.

\bibliography{refs}
\bibliographystyle{splncs04}

\newpage
\appendix
\renewcommand*{\theHsection}{appendix.\arabic{section}}
\renewcommand*{\theHsubsection}{appendix.\arabic{section}.\arabic{subsection}}

This document provides supplementary details for the main paper. We include dataset statistics and provenance (Sec.~\ref{sec:datasets}), pseudocode for TITAnD, CFT, and QuadTree (Sec.~\ref{sec:pseudocode}), encoder implementation details (Sec.~\ref{sec:encoder_details}), per-benchmark training hyperparameters (Sec.~\ref{sec:training_details}), robustness and anomaly-type analyses (Sec.~\ref{sec:robustness}), extended qualitative analysis (Sec.~\ref{sec:qualitative}), and the relation to period-folding time-series models (Sec.~\ref{sec:period_folding}).

\section{Dataset Details}
\label{sec:datasets}

\begin{table}[h]
\caption{Dataset statistics.}
\label{tab:dataset}
\centering
\renewcommand{\arraystretch}{1.1}
\resizebox{\textwidth}{!}{%
\begin{tabular}{l ccc ccc cc}
\toprule
 & \multicolumn{3}{c}{\textbf{Scale}} & \multicolumn{3}{c}{\textbf{Splits}} & \multicolumn{2}{c}{\textbf{Anomaly Prevalence}} \\
\cmidrule(lr){2-4} \cmidrule(lr){5-7} \cmidrule(lr){8-9}
Benchmark & Modality & Agents & Duration & Train & Val & Slots/Agent & Slot Rate & Agent Rate \\
\midrule
Sparse (NumoSim-LA) & Stay-points & 80,381 & 55 days & 50,238 & 30,143 & 15,840 & 0.024\% & 0.47\% \\
Dense (Tokyo) & GPS (10\,s) & 18,469 & 66 days & 14,772 & 3,697 & 6,336 & 0.20\% & 46.4\% \\
Dense2Sparse (Tokyo) & Stay-points & 80,463 & 120 days & 50,416 & 30,047 & 34,560 & 0.009\% & 0.57\% \\
Sparse (PoL, Atlanta) & Stay-points & 3,000 & 464 days & 2,400 & 600 & 133,632 & --- & 5.0\% \\
\bottomrule
\end{tabular}
}%
\end{table}

Slot-level anomalies are 20--230$\times$ rarer than anomalous agents on every benchmark with slot labels (Table~\ref{tab:dataset}). Because the chance level of AUC-PR equals the positive rate, temporal AUC-PR is lower than agent AUC-PR for every model in Table~\ref{tab:main_results}, and the two granularities should not be compared on raw values. For the same reason, temporal mIoU stays near 0.5 even for weak detectors: the normal-class IoU is close to 1 when almost every slot is normal.

\textbf{Sparse (NumoSim-LA).}
NumoSim~\cite{shafqat2024numosim} provides natively sparse stay-point trajectories from the Los Angeles region.
Each event records arrival time, departure time, and a POI location; no dense GPS is available.
Trajectories span ${\sim}$55 days (28 past + 27 future).
Anomaly labels are aggregated to 5-minute slots via the sparse labeling rule of Sec.~\ref{sec:overview}; a trip slot takes the trip's own label where the benchmark provides one (Dense2Sparse) and otherwise the label of either adjacent stay.

\textbf{Dense (Tokyo).}
A private dense GPS dataset from the Tokyo metropolitan region sampled at ${\sim}$10-second resolution.
Each agent spans 66 days (33 past + 33 future), yielding ${\sim}$570K GPS points per agent.
Anomaly labels are provided at per-point granularity and aggregated to 15-minute slots via OR aggregation (Sec.~\ref{sec:overview}).
The trajectories come from an agent-based patterns-of-life simulator. Synthetic agents are sampled from public census microdata, activity choices are produced by a random-utility model calibrated on household-travel and time-use surveys, and mobility is realized on an OpenStreetMap road network using map-constrained routing. An independent red team injected anomalies without disclosing the rules to the modeling team.

\textbf{Dense2Sparse (Tokyo).}
A larger, distinct cohort from the same Tokyo source is converted to stay-point sequences spanning 120 days. The shared geography and collection process permit a modality comparison, but the different agents and horizons mean this is not a paired dense-versus-sparse ablation.

\textbf{Patterns of Life (PoL).}
PoL~\cite{zufle2023urbanlife,amiri2023pol,rb_amiri24} is a public sparse benchmark simulated on a map of Atlanta, with 3,000 agents over a 450-day history and a 14-day evaluation window. Its release identifies 150 anomalous agents over a 14-day evaluation window but does not provide per-slot anomaly labels for that window, so we report agent AUC-PR rather than temporal localization metrics. Because this leaves only 30 anomalous validation agents, agent AUC-PR varies across training runs; for TITAnD and SegFormer, the two strongest methods, PoL entries are the mean over six training seeds. Each stay-point pixel additionally encodes the category of the visited venue (apartment, workplace, restaurant, or pub), the semantic information also available to LM-TAD, and stay-points are split at calendar-day boundaries. Because PoL labels agents rather than stays, TITAnD is trained with weak slot targets: every stay of an anomalous agent within the 14-day window is positive, trip slots inherit the label of their adjacent stays, and the loss covers that window only. Table~\ref{tab:hyperparams} lists the remaining settings. The supervised LM-TAD adaptation uses place-token BCE with positive class weight 70, learning rate $3\times10^{-4}$, and batches of 256 daily sequences. DINOv2+Adapter uses the same PoL input as TITAnD.

\section{Pseudocode}
\label{sec:pseudocode}

\subsection{TITAnD: End-to-End Forward Pass}

Listing~\ref{lst:titand} shows the high-level three-stage pipeline. Encoder internals (spatial encoding, temporal calendar encoding, kinematic features) are detailed in Sec.~\ref{sec:tokenizer_method}; the pseudocode focuses on inter-stage data flow.

\begin{lstlisting}[style=pseudopython, caption={TITAnD forward pass.}, label={lst:titand}]
def titand_forward(batch):
    B, D, S = batch.mask.shape

    # Stage 1: Modality-specific encoder
    # (DenseTrajEmbed or SparseTrajEmbed)
    sp = encode_spatial(batch)     # [B,D,S,128]
    tm = encode_temporal(batch)    # [B,D,S,66]
    kn = encode_kinematic(batch)   # [B,D,S,32]
    x = mlp_fuse(cat(sp, tm, kn)) # [B,D,S,C]
    x += type_emb(batch.stop_frac)
    x += venue_emb(batch.venue)  # PoL

    # Stage 2: CFT
    x = cft(x)

    # Stage 3: Prediction head
    y = mlp_clf(x).squeeze(-1)    # [B,D,S]
    y_agent = y[observed].max()   # [B]
    return y, y_agent
\end{lstlisting}

\subsection{Cyclic Factorized Transformer (CFT)}

Listing~\ref{lst:cft} details the interleaved block structure described in Sec.~\ref{sec:cft_method}. Each block applies intra-day attention (over $S$ slots within each day) followed by inter-day attention (over $D$ days at the same time-of-day), both with LayerNorm and residual connections.

\begin{lstlisting}[style=pseudopython, caption={CFT forward pass.}, label={lst:cft}]
def cft_forward(x):
    B, D, S, C = x.shape
    # position encoding, once on the input (as in ViT)
    x = rope_slot(x)  # slot position
    x = rope_day(x)   # day position

    for i in range(L):
        # Intra-day: self-attn over S per day
        x = x + intra_attention[i](layer_norm(x))
        x = x + ffn_intra[i](layer_norm(x))

        # Inter-day: self-attn over D per slot
        x = x + inter_attention[i](layer_norm(x))
        x = x + ffn_inter[i](layer_norm(x))
    return x # [B, D, S, C]
\end{lstlisting}

\subsection{QuadTree Spatial Discretization}

Listing~\ref{lst:quadtree} shows the adaptive grid construction referenced in Sec.~\ref{sec:tokenizer_method}. Cells with more than \texttt{cap} POIs are recursively split into quadrants until reaching the minimum resolution.

\begin{lstlisting}[style=pseudopython, caption={QuadTree construction.}, label={lst:quadtree}]
def build_quadtree(bounds, pois,
        base=200, cap=125, min_sz=10):
    # base: grid pitch (m) for (i,j)
    # cap: POI split threshold
    # min_sz: smallest cell (m)
    root = Node(bounds)
    queue = [(root, all_pois)]
    leaves = []

    while queue:
        node, idx = queue.popleft()
        half = min(node.w, node.h) / 2

        if len(idx) > cap and half >= min_sz:
            for c in split_quadrants(node):
                c_idx = idx[in_bounds(pois, c)]
                queue.append((c, c_idx))
        else:
            leaves.append(node)

    for i, leaf in enumerate(leaves):
        leaf.cell_id = i
    lut = build_lookup(leaves, res=5)
    return leaves, lut
\end{lstlisting}

\subsection{Spatio-Semantic Cell Embedding (\texorpdfstring{$\mathrm{Enc}_{\text{sp}}$}{Enc\_sp})}

Listing~\ref{lst:encsp} shows the shared \texttt{cell\_embed} core (GPS $\to$ QuadTree $\to$ 128-d).
The dense encoder refines each of the $P{=}90$ cell embeddings in a slot with the point's offset inside its cell and pools them via learned attention.
The sparse encoder embeds stops by the cell at their location; trip slots carry no spatial feature and take their spatial context from neighbouring stops through attention.

\begin{lstlisting}[style=pseudopython, caption={Spatio-semantic cell embedding.}, label={lst:encsp}]
def cell_embed(cell_id, i, j):
    # Semantic: attention-pool POIs in cell
    h = poi_mlp(cat(cat_emb(poi_cat[cell_id]),
                    pos_mlp(dxy[cell_id]))) # [N,64]
    alpha = softmax(w_sem @ h, poi_mask)
    s = dot(alpha, h) if n_poi[cell_id] else e_empty # [64]
    # e_empty: learned embedding for cells without POIs
    # Spatial: sinusoidal grid PE
    pe = sinusoidal_pe(i, j)  # [128]
    return mlp_sp(cat(s, pe)) # [128]

# Dense: pool P cell embeddings per slot
def enc_sp_dense(gps):        # [P, 2]
    cells, ij, off = quadtree(gps)
    e = vmap(cell_embed)(cells, ij) # [P, 128]
    e = refine(e, off)        # point offset in cell
    beta = softmax(v_pool @ tanh(W_pool @ e))
    return dot(beta, e)       # [128]

# Sparse stop: cell embedding at location
def enc_sp_stop(stop):
    cell, ij = quadtree(stop.gps)
    return cell_embed(cell, *ij)    # [128]

# Sparse trip: no spatial feature
def enc_sp_trip(trip):
    return zeros(128)               # [128]
\end{lstlisting}

\subsection{Temporal Encoder (\texorpdfstring{$\mathrm{Enc}_{\text{t}}$}{Enc\_t})}

Listing~\ref{lst:enct} shows the shared temporal encoder used identically for both dense and sparse encoders.
Six calendar attributes are encoded independently and concatenated to a 66-d vector: hour, day of week, day of month, and month use fixed sinusoidal encodings, while part of day and week of month use learned embeddings.

\begin{lstlisting}[style=pseudopython, caption={Temporal encoder.}, label={lst:enct}]
def enc_temporal(t):
    # 6 calendar attributes -> 66-d total
    h = cat(
        sincos(t.hour, 24),       # 8-d
        sincos(t.day_of_week, 7), # 6-d
        sincos(t.day_of_month, 31), # 8-d
        sincos(t.month, 12),      # 12-d
        emb_pod(t.part_of_day),   # 16-d, learned
        emb_wom(t.week_of_month), # 16-d, learned
    )
    return h                      # [66]
\end{lstlisting}

\subsection{Kinematic Encoders (\texorpdfstring{$\mathrm{Enc}_{\text{kn}}$}{Enc\_kn})}

Listing~\ref{lst:enckn} shows both variants.
The dense encoder derives an 11-d descriptor from raw GPS within a 15-minute slot.
The sparse encoder uses a shared 2-d input for stops and trips: stops set distance to zero (signalling a stationary event), while trips carry travel duration and distance.
Stops and trips share one linear projection; the dense encoder uses its own two-layer MLP.

\begin{lstlisting}[style=pseudopython, caption={Kinematic encoders (dense and sparse).}, label={lst:enckn}]
# Dense: 11-d from ~90 GPS points per slot
def enc_kn_dense(xy, t):  # [P, 2] metres, [P] s
    v = diff(xy) / diff(t)        # [P-1, 2]
    spd = norm(v, dim=-1)         # [P-1]
    brg = atan2(v[:,0], v[:,1])   # [P-1]
    ms, mc = sin(brg).mean(), cos(brg).mean()
    b = atan2(ms, mc)             # circular mean
    acc = norm(diff(v), dim=-1) / diff(t)[:-1]
    f = cat([v.mean(0)/200, sin(b), cos(b),
             spd.mean()/200, 1 - hypot(ms, mc),
             acc.mean()/50,
             spd.min()/200, spd.max()/200,
             acc.min()/50, acc.max()/50]) # 11-d
    return mlp_kn(f)  # Linear-ReLU-Linear, [32]

# Sparse stop: duration only, distance = 0
def enc_kn_stop(stop):
    f = [log10(1+stop.dur), 0]  # 2-d
    return gelu(linear_kn(f))   # [32]

# Sparse trip: duration and distance
def enc_kn_trip(trip):
    f = [log10(1+trip.dur),
         log10(1+trip.dist)]    # 2-d
    return gelu(linear_kn(f))   # [32]
# linear_kn shared across stops and trips
\end{lstlisting}

\section{Encoder Implementation Details}
\label{sec:encoder_details}

\subsection{QuadTree Grid Statistics}

The QuadTree starts from the bounding box of the region (the AOI polygon for Tokyo, the buffered POI extent for NumoSim-LA and PoL) and recursively splits any cell containing more than 125 POIs into four quadrants, stopping when children would be smaller than 10\,m (Tokyo and NumoSim-LA; PoL, a 4\,km area, uses 100\,m base cells and a 25-venue threshold).
This yields fine resolution in dense commercial districts and coarse cells in sparse residential areas.

\begin{table}[h]
\centering
\small
\begin{tabular}{lc}
\toprule
Property & Value \\
\midrule
Total leaf cells (Tokyo) & ${\sim}$78,000 \\
Split threshold & 125 POIs \\
Minimum cell size & 10\,m \\
Lookup resolution & 5\,m (for $O(1)$ vectorized queries) \\
\bottomrule
\end{tabular}
\end{table}

\subsection{POI Categories}

For Tokyo we use 45 semantic POI categories grouped from raw OpenStreetMap tags (NumoSim-LA uses 16, and PoL its 4 venue categories):

\begin{table}[h]
\centering
\small
\begin{tabular}{ll}
\toprule
Category group & Examples \\
\midrule
Residential (6) & apartment, house, dormitory, \ldots \\
Commercial (8) & store, mall, supermarket, \ldots \\
Food \& Drink (5) & restaurant, cafe, bar, \ldots \\
Office (4) & office building, coworking, \ldots \\
Healthcare (4) & hospital, clinic, pharmacy, \ldots \\
Education (5) & school, university, library, \ldots \\
Transportation (4) & station, parking, bus stop, \ldots \\
Recreation (5) & park, gym, cinema, \ldots \\
Other (4) & religious, industrial, \ldots \\
\bottomrule
\end{tabular}
\end{table}

\subsection{Feature Dimensions}

\textbf{Spatio-semantic cell embedding.}
Each POI is encoded by its category embedding (32-d) concatenated with a relative position MLP (32-d), yielding a 64-d POI feature.
Attention pooling over all POIs in a cell produces a 64-d cell summary, which is concatenated with a 128-d sinusoidal grid position encoding PE$(i,j)$ and projected via $\mathrm{MLP}_{\text{sp}}$ to a 128-d cell embedding.
For the dense encoder, learned attention pooling over the $P{=}90$ refined cell embeddings per slot produces a 128-d spatial token.

\textbf{Temporal encoder.}
Six calendar attributes are independently encoded and concatenated:
hour-of-day (8-d), day-of-week (6-d), day-of-month (8-d), month-of-year (12-d), part-of-day with 6 four-hour bins (16-d), and week-of-month (16-d), totaling 66 dimensions.

\textbf{Kinematic encoder.}
For the dense encoder, $\mathrm{Enc}_{\text{kn}}$ takes an 11-d descriptor (velocity components, direction, speed, turning variability, acceleration, min/max speed and acceleration) and projects it to 32 dimensions.
For the sparse encoder, $\mathrm{Enc}'_{\text{kn}}$ takes a 2-d input (log-duration, log-distance) and projects it to 32 dimensions.

\textbf{Slot fusion.}
The three feature groups are concatenated ($128 + 66 + 32 = 226$ dimensions) and projected to $C{=}256$, by a linear layer with layer normalization in the sparse encoder and by a two-layer MLP in the dense encoder.

\section{Training Details}
\label{sec:training_details}

All benchmarks use a positively weighted binary cross-entropy (plus a Dice term for the benchmarks with $\lambda{>}0$ in Table~\ref{tab:hyperparams}), a linear learning-rate warmup followed by cosine decay, gradient clipping at 1.0, mixed precision, and SpecAugment with probability 0.5. Table~\ref{tab:hyperparams} lists the main settings that differ across benchmarks. For every method, Table~\ref{tab:main_results} reports the checkpoint with the best validation score. The dense CFT has 16.2M parameters.

\begin{table}[h]
\caption{Per-benchmark training hyperparameters. $^\dagger$Muon~\cite{liu2025muon} for weight matrices and AdamW for the remaining parameters.}
\label{tab:hyperparams}
\centering
\small
\setlength{\tabcolsep}{3pt}
\renewcommand{\arraystretch}{1.1}
\begin{tabular}{lcccc}
\toprule
Hyperparameter & \shortstack{Dense\\(Tokyo)} & \shortstack{Sparse\\(NumoSim-LA)} & \shortstack{Dense2Sparse\\(Tokyo)} & PoL \\
\midrule
Optimizer & Muon$^\dagger$ & AdamW & AdamW & AdamW \\
Learning rate & 5$\times$10$^{-3}$ & 7$\times$10$^{-5}$ & 2$\times$10$^{-4}$ & 2$\times$10$^{-4}$ \\
LR warmup epochs & 10 & 5 & 5 & 5 \\
Epochs & 40 & 20 & 20 & 20 \\
Batch size per GPU & 32 & 64 & 32 & 8 \\
Gradient accumulation & 1 & 1 & 1 & 8 \\
Weight decay & 0.1 & 0.01 & 0.01 & 0.01 \\
Positive-class weight & 400 & 70 & 70 & 5 \\
Dice weight $\lambda$ & 0 & 1.0 & 0 & 0.3 \\
Label smoothing & 0.02 & 0 & 0 & 0 \\
\bottomrule
\end{tabular}
\end{table}

The comparison methods share the data splits, labels, and metrics. U-Net and SegFormer start from ImageNet-pretrained encoders and DINOv2+Adapter from a pretrained DINOv2 backbone, each fine-tuned end-to-end; CFT and the flat Transformer are trained from scratch. Chronos-2 uses the Chronos-2 (small) architecture trained from scratch, with a per-slot classification head in place of its forecasting head; on Dense Tokyo it reads coordinates and POI channels at 5-minute resolution. USTAD, ICAD, and LM-TAD are trained from scratch as supervised variants of their published architectures.

\section{Robustness and Anomaly-Type Analysis}
\label{sec:robustness}

\textbf{NumoSim anomaly types.}
NumoSim-LA distinguishes unfamiliar-location visits from familiar-location revisits at shifted times. TITAnD obtains 0.51 temporal AUC-PR on unfamiliar-location anomalies and 0.50 on shifted-time revisits. The two are detected at nearly the same rate even though shifted-time revisits are spatially and semantically familiar, leaving timing relative to the agent's routine as the only signal. Tokyo does not expose anomaly-type labels because the injection scheme was withheld by the red team, so the same quantitative stratification is not available there.

\textbf{Routine drift.}
HTI explicitly retains month and time-of-day features, while its semantic channels can map changed physical locations to related activity categories. These features may help with timing or location drift, but the present benchmarks do not provide a controlled drift annotation. We therefore treat robustness to seasonal schedule changes, relocation, and evolving routines as an open evaluation question rather than a demonstrated guarantee.

\section{Extended Qualitative Analysis}
\label{sec:qualitative}

Figure~\ref{fig:qualitative} compares the ground truth, CFT predictions, and attention rollout for four Dense (Tokyo) agents. Attention is averaged over heads and layers; inter-day attention is additionally averaged over anomalous time slots, while intra-day attention is averaged over anomalous days. Across cases, inter-day attention concentrates on anomalous days and intra-day attention highlights the corresponding within-day windows.

For each agent, we compare the anomaly period against the established pattern of life during the 33-day baseline (Days 0--32).
Throughout this analysis, \textit{unique locations} refers to distinct grid cells visited; \textit{novel locations} are cells not visited during the baseline; \textit{POI enrichment} is the ratio of POI category exposure during anomaly versus baseline; and \textit{velocity} is normalized per agent using min-max scaling.

\subsection{Agent 1 (Day 41, 18:00--22:00)}

During the baseline period, this agent follows a compact routine with limited variation. On Day 41 evening, the behavior shifts abruptly: the agent traverses a substantially wider region and exhibits different stop-and-go dynamics within a four-hour window. Inter-day attention peaks on Day 41, while intra-day attention concentrates on the ground-truth interval. The window combines a surge in distinct visited locations with extended stationary periods at atypical areas.

\subsection{Agent 3 (Day 37, 09:00--13:00)}

This anomaly is not dominated by spatial displacement. On Day 37 morning, the agent is stationary for most of the four-hour interval and exhibits repeated prolonged dwells, deviating from the baseline weekday pattern of regular movement and brief stops. Inter-day attention isolates Day 37 and intra-day attention concentrates on 09:00--13:00, showing how fine-grained temporal dynamics can support localization even when the visited region remains similar.

\subsection{Agent 2 (Days 40--41, 00:00--23:45)}

This agent exhibited the longest anomaly in our evaluation: 24 continuous hours.
During baseline, the agent frequented only 37 unique locations with strong residential preference (66\%) and essentially zero engagement with entertainment venues ($<$0.5\%).
Starting midnight on Day 40, the agent visited 86 locations (83\% novel) with a dramatic semantic shift: 6.4$\times$ enrichment in night club proximity and 2.4$\times$ enrichment near theaters---POI categories absent from their pattern of life.
Between 21:15--22:30 on Day 40, the agent made five consecutive 15-minute stops in unpopulated areas, behavior inconsistent with normal evening activity.
The 24-hour duration, spatial novelty, and entertainment venue engagement collectively constitute a comprehensive behavioral discontinuity.

\subsection{Agent 4 (Day 50, 06:00--10:00)}

This case presents the most geographically dramatic anomaly.
Baseline morning activity showed a consistent 5\,km commute with 2--3 brief stops.
On Day 50, the agent traversed a 31$\times$14\,km area---spanning the metropolitan region north-to-south---visiting 267 locations (81\% novel).
The trajectory formed an approximately linear path, suggesting deliberate transit rather than exploration.
Most notably, the five longest dwells (15 minutes each, 07:00--08:15) occurred at education and religious facilities---institutional POI categories with negligible baseline presence.
Despite covering 6$\times$ normal distance, velocity decreased 81\%, indicating methodical movement with frequent stops.
This combination of extreme spatial displacement, institutional engagement, and paradoxically reduced speed triggered high-confidence detection.

\section{Relation to Period-Folding Time-Series Models}
\label{sec:period_folding}
TimesNet~\cite{wu2023timesnet} and TimeMixer++~\cite{timemixer++} fold a regularly sampled multivariate series into a period $\times$ phase grid whose period is estimated with the Fourier transform, and model within- and across-period variation with 2D convolutions (TimesNet) or axis-wise attention (TimeMixer++). TITAnD shares the idea of a two-dimensional view of time, but differs in motivation, data, representation, architecture, and task.

\textbf{Role of the two-dimensional view.} In TimesNet and TimeMixer++, the 2D view is a feature transformation applied inside a general-purpose backbone, with periods discovered from each signal; TimesNet re-estimates them at every layer. In TITAnD, the image is the task formulation. It maps dense and sparse trajectories into one representation, so the same architecture serves both; it compresses months of observations into a grid of $D\times S$ slots on which attention is factorized along the two axes, reducing cost from $O(D^2S^2C)$ to $O((DS^2{+}SD^2)C)$; and it aligns those axes with within-day and across-day human routines, so slots within a day capture episodes and days at the same time of day capture routines.

\textbf{Data.} These models take a single regularly sampled series. TITAnD takes trajectories in two forms, dense GPS streams and sparse stay-point events, and maps both into the same image.

\textbf{Multi-modal pixels.} In these models, a grid cell holds a projection of the measured values at one time step. In HTI, a pixel fuses spatial (map cell), semantic (nearby points of interest), temporal (calendar), and kinematic (speed, acceleration, bearing, and turning on dense GPS; duration and distance on sparse events) information aggregated from all observations inside its time-of-day slot.

\textbf{Architecture that exploits the structure.} The image is designed together with its backbone. CFT applies intra-day attention over the slots of each day to model episodes and inter-day attention across days at the same time of day to model routines, with separate rotary position encodings per axis. TimeMixer++ also uses axis-wise attention, but over estimated periods to separate seasonal and trend components; TimesNet uses local 2D convolutions, the operator class evaluated on HTI through the U-Net backbone in Table~\ref{tab:main_results}, where CFT outperforms it. Sec.~\ref{sec:exp_ablation} further shows that neither axis alone recovers full performance.

\textbf{Task.} TITAnD is trained with supervision for per-slot anomaly localization and agent-level detection over multi-month horizons of tens of thousands of cells, whereas TimesNet detects abnormal time points by reconstructing 100-step windows of sensor readings.
\end{document}